\pdfoutput=1
\documentclass[letterpaper, 10 pt, conference]{ieeeconf}  

\IEEEoverridecommandlockouts                              

\usepackage{graphicx}
\usepackage{epsfig} 
\usepackage{mathptmx} 

\usepackage{times} 
\usepackage{amsmath} 
\usepackage{amssymb}  
\usepackage{mathtools}
\usepackage{xcolor}
\usepackage{eucal}
\usepackage{indentfirst}
\usepackage{algorithm,algorithmic}
\usepackage{physics}
\usepackage{multirow}
\usepackage{multicol}

\usepackage{subcaption}
\usepackage{bm}
\newcommand{\R}{\mathbb{R}}

\newcommand{\INDSTATE}[1][1]{\STATE\hspace{#1\algorithmicindent}}

\begin{document}
\title{\LARGE \bf
Hybrid LMC: Hybrid Learning and Model-based Control for Wheeled Humanoid Robot via Ensemble Deep Reinforcement Learning
}

\author{Donghoon Baek$^{1}$, Amartya Purushottam$^{2}$, and Joao Ramos$^{1,2}$
\thanks{This work is supported by the National Science Foundation via grant IIS-2024775.}
\thanks{The authors are with the $^1$ Department of Mechanical Science and Engineering and the $^2$ Department of Electrical and Computer Engineering at the University of Illinois at Urbana-Champaign, USA.{\tt\small jlramos@illinois.edu}} 
}

\maketitle

\begin{abstract}
Control of wheeled humanoid locomotion is a challenging problem due to the nonlinear dynamics and underactuated characteristics of these robots.
Traditionally, feedback controllers have been utilized for stabilization and locomotion. However, these methods are often limited by the fidelity of the underlying model used, choice of controller, and environmental variables considered (surface type, ground inclination, etc).
Recent advances in reinforcement learning (RL) offer promising methods to tackle some of these conventional feedback controller issues, but require large amounts of interaction data to learn.
Here, we propose a hybrid learning and model-based controller \emph{Hybrid LMC} that combines the strengths of a classical linear quadratic regulator (LQR) and ensemble deep reinforcement learning. Ensemble deep reinforcement learning is composed of multiple Soft Actor-Critic (SAC) and is utilized in reducing the variance of RL networks. By using a feedback controller in tandem the network exhibits stable performance in the early stages of training. As a preliminary step, we explore the viability of \emph{Hybrid LMC} in controlling wheeled locomotion of a humanoid robot over a set of different physical parameters in MuJoCo simulator. Our results show that \emph{Hybrid LMC} achieves better performance compared to other existing techniques and has increased sample efficiency.

\end{abstract}

\section{Introduction}
\label{S:1}
Humanoid robots have the potential to aid workers in physically demanding and dangerous jobs such as firefighting and disaster relief \cite{wang2015hermes,ramos2019dynamic}. In order to aid in these tasks, humanoid robots must be capable of manipulation and locomotion, while being robust to intermittent contact and disturbances. Wheeled-humanoid robots (WHR) are emerging as promising platforms for accomplishing these tasks by combining advantages of mobile robots with the dexterity of legged robots \cite{klemm2019ascento, li2019wlr}.

However, inherent instability, nonlinearity, inaccurate modeling error, and strongly coupled mechanism pose challenges to control WHR. Specifically, balancing control of the WHR is a pivotal role for the robots to transverse various terrains in the real world.

The most common approach of control for these high dimensional nonlinear systems is to model a robot using reduced-order models (RoMs), such as Linear Inverted Pendulums (LIP) and Wheeled Inverted Pendulums (WIP), and adopt model-based linear quadratic regulator (LQR) \cite{klemm2020lqr, xin2020online} or model predictive control (MPC) \cite{onkol2018adaptive}. Alternatively, differential  dynamic programming (DDP) and Nonlinear MPC 

\begin{figure}[t]
\begin{center}
\includegraphics[width=0.7\linewidth]{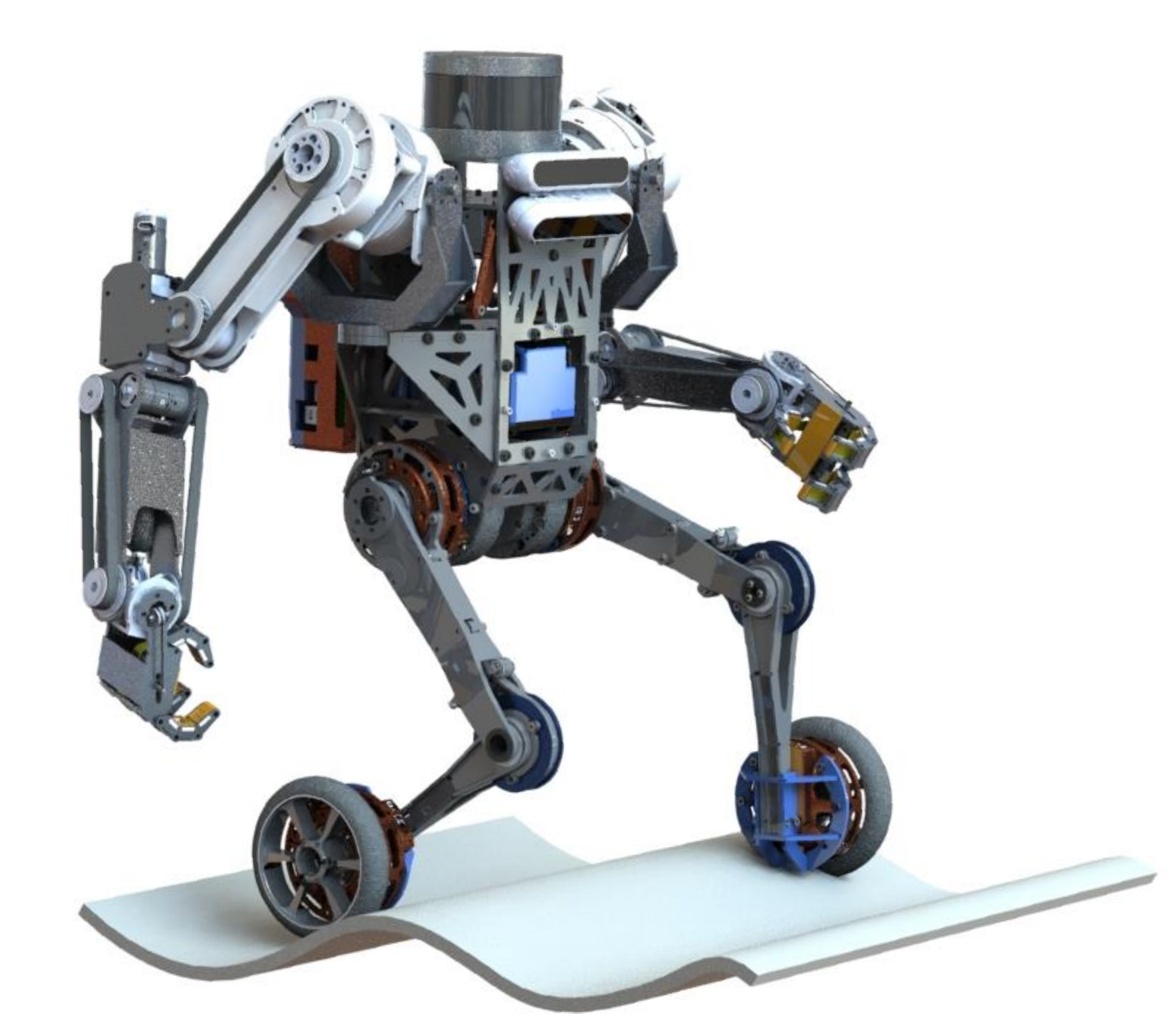}
\end{center}
\caption{Wheeled Humanoid Robot System, SATYRR.}
\label{fig1}
\end{figure}

 \noindent (NMPC) are utilized to generate whole-body motion as a nonlinear approach \cite{zafar2019hierarchical,grandia2020nonlinear}. Despite their widespread usage among the robotic community, the stability and robustness of these controllers are limited by the fidelity of the robot model and of the surrounding environment. Besides, the performance of these methods depends on the accuracy of the model which has an inherent error.

Deep reinforcement learning (RL)-based methods have garnered a growing amount of attention recently as an up-and-coming solution and have shown the success of tackling highly nonlinear locomotion problems \cite{cui2021learning, haarnoja2018learning, hwangbo2019learning}. They can overcome the limitations of prior model-based approaches by learning a policy directly from experience and automatically tuning the controller to optimize the given reward (or cost) function representing the task. However, standard RL methods require long interaction between the robot and an environment to learn complicated skills, which can be unsafe initially. Collecting the amount of data that is needed to learn a complex task is time-consuming. Although many Sim-to-Real techniques are suggested \cite{hwangbo2019learning, kumar2021rma, lee2020learning}, reducing the domain gap between a simulation and reality is still challenging and takes an extensive amount of time, up to several days, to train. Exceptionally, control of WHRs solely with RL is challenging since they are inherently unstable at the initial stage during exploration, and re-setup of the robot every time is significantly inefficient and risky.  



Meanwhile, the incorporation of the inductive bias or prior knowledge (e.g., analytical model, a conventional controller) with RL looks to address the issues of a conventional controller and RL-based methods by aiding the RL policy to be explored more safely and fast through increasing sample 

\begin{figure*}[t]
\centerline{\includegraphics[width=16cm]{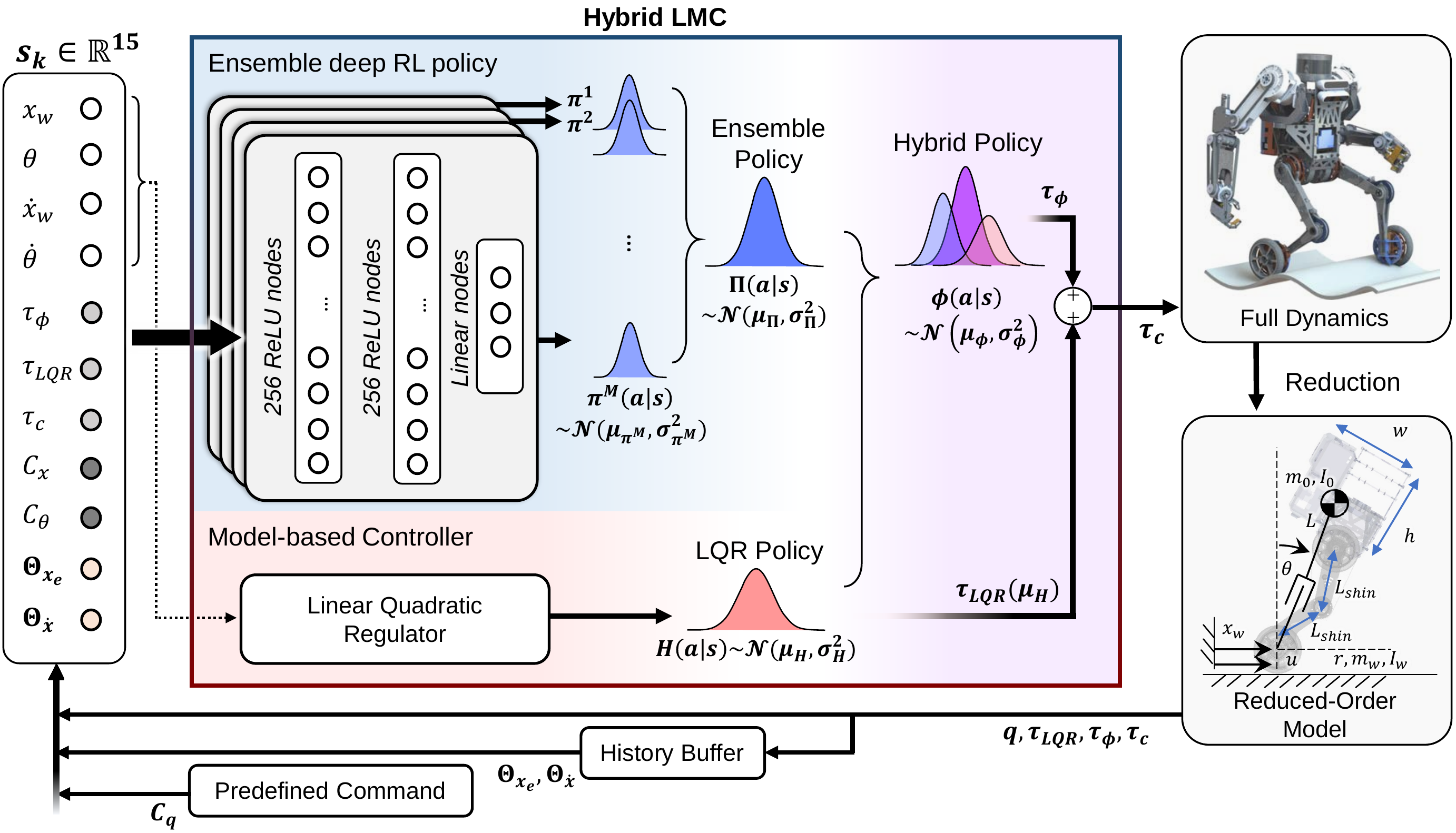}}
    \caption{\textbf{Overall Pipeline of Hybrid LMC.} \emph{Hybrid LMC} generates the compensated torque $\tau_c$ directly (End-to-End) via a combination of the hybrid policy $\Pi(a|\bm{s})$ and a LQR controller $H(a|\bm{s})$. Hybrid policy $\Pi(a|\bm{s})$ is obtained by using an ensemble deep RL policy that is a mixture of a single SAC policy $\pi(a|\bm{s})$ with $\bm{s} \in \R^{15}$ at time $k$ as an input. State $\bm{s}$ comprises the states of the WIP model of the robot as well as other augmented states. All states are defined in the section \ref{experimetB}.}
    \label{fig2}
\end{figure*}

\noindent efficiency and reducing state space volume \cite{johannink2019residual,rana2020multiplicative,rana2021bayesian}. Although this approach has shown impressive results on manipulation and navigation tasks \cite{johannink2019residual,rana2020multiplicative}, it has yet to be shown for locomotion that is a high dimensional and  challenging to collect task-relevant data. Specifically, most controllers for WHRs map from the command to the resulting torque straightly without using a high-level trajectory or a learned policy, and this results in addressing the locomotion problem more challenging.\\
\indent With this in mind, the goal of this paper is to develop a hybrid controller for a WHR that complements each of the RL and a model-based controller by starting exploration from a relatively stable controller as well as effectively reducing a residual error of the control part.\\
\indent In this work, a hybrid learning and model-based controller (\emph{Hybrid-LMC}) combining an optimal controller and ensemble deep reinforcement learning is proposed to enhance the locomotion control performance by reducing a residual error resulting from nonlinearities, modeling error, and a variety of environmental changes. The fundamental concept is the same with the residual reinforcement learning \cite{johannink2019residual}, but unlike the previous work, we utilized the ensemble RL policy that leverages multiple Soft Actor-Critic (SAC) \cite{haarnoja2018soft} and distributional action torque provided by an optimal controller (LQR) to choose the compensated torque more carefully through broader exploration with low variance. Our approach allows generating a torque directly contrary to the existing works that use a policy network to build a high-level command such as a trajectory signal. \\
\indent The contributions of this work are as follows: (1) Hybrid learning and model-based controller, taking the advantage of both a model-based controller and a deep reinforcement learning for increasing the control performance of wheeled-legged humanoid robots, is proposed. To the best of our knowledge, this is the first trial to apply the combined policy 

\begin{table}[t] \caption{SATYRR specification.}
\centering 
\label{t_specification}
\setlength{\tabcolsep}{2pt}
\begin{tabular}{|c|c|c|c|c|c|}
\hline
Parameter & Value      & Parameter & Value                      & Parameter & Value    \\ \hline
$m_0$      & $6.8kg$      & $I_0$      & $0.16kgm^2$ & $m_w$      & $0.4297kg$ \\ \hline
$L$         & $0.28m$      & $r$         & $0.06m$                      & $h$         & $0.26m$    \\ \hline
$I_w$      & $0.00278kgm^2$ & $L_{sine}$     & $0.15m$                      & $w$         & $0.22m$    \\ \hline
\end{tabular}
\end{table}
\noindent  such as residual RL to humanoid locomotion. (2) Experimental results indicate that \emph{Hybrid LMC} outperforms residual reinforcement learning and model-free reinforcement learning algorithms as well as compensates the residual error of a LQR controller even in the situation where the diverse physical parameter has changed. (3) Ablation study and additional experiments for investigating \emph{Hybrid LMC} utilizing efficiently are carried out and analyzed carefully. (4) Experimental result using a human data shows the feasibility of \emph{Hybrid LMC} applying to a teleoperated system.

\section{Method}
\label{method}

In this section, we discuss the wheeled-legged humanoid robot platform of our choice, SATYRR, its RoM and LQR controller, and the proposed \emph{Hybrid LMC} that combines an ensemble deep RL and LQR. The \emph{Hybrid LMC} pipeline can be seen in Fig. \ref{fig2}. We explain its details in section \ref{methodB}



\subsection{Modeling and Feedback Control}
SATYRR (Fig. \ref{fig1}) is an anthropomorphic biped robot with two powered wheels in place of its feet. We describe the main parameters of the SAYTRR model in Table. \ref{t_specification}.      

 Here we use the WIP  \cite{klemm2020lqr,chen2020underactuated} RoM model that consists of the wheels and a lumped rigid body that represents the robots torso as seen in Fig. \ref{fig2}. The dynamics of this model are given by: 

\begin{equation}\label{wip_dyn}
    \begin{gathered}[b] 
        \bigg (m_{o} \!+\! m_{w} \!+\! \frac{I_w}{r^2}\bigg )\ddot{x}_w \!+\! m_{o}L s_{\theta}\dot{\theta}^2 
    \!-\! m_{o}L c_{\theta}\ddot{\theta} \!= \!u\\
        (m_{o}L^2 + I_{o})\ddot{\theta} - m_{o}Lc_{\theta}\ddot{x}_w - m_{o} gLs_{\theta} \!= \!0 
    \end{gathered}\
\end{equation}



\noindent where $x_w$ denotes the traversed position of SATYRR calculated as an average of two wheel encoders, $\theta$ is the pitch angle. $\dot{x}_w$ and $\dot{\theta}$ represent their corresponding derivatives. The mass of the body and wheel indicates $m_{0}$ and $m_w$, respectively. $u$ is the control input and torque applied to the wheel, $r$ is the radius of the wheel, and $g$ is gravity. The length between the center of the wheel and the center of mass (CoM) of the body is denoted by $L$, $I_w$ is the inertia of the wheel, and $I_{0}$ is the inertia of the body. 

Defining the state vector $\bm{q} = [x_w \quad \theta \quad \dot{x}_w \quad \dot{\theta}]^\top$, we linearize \ref{wip_dyn} around the upright equilibrium to obtain the state space equations and resulting optimal gains for the LQR: $\bm{K} = [-100, -315, -40, -40]$. For regulating yaw motion and the height of the robot conventional PD controller are used. 

\color{black}
\subsection{Hybrid Learning and Model-based Controller}
\label{methodB}
\emph{Hybrid LMC} shares fundamental concepts with residual RL \cite{johannink2019residual} in that they combine a conventional feedback controller with a learned RL policy. In this manner, the two controllers complement each other and compensate for their individual shortcomings: 
\begin{equation}
\tau_c = \tau_{LQR}(\bm{s}) + \tau_{\phi}(\bm{s})
\label{eqn1}
\end{equation}


\noindent where $\tau_{LQR}(\bm{s})$ and $\tau_{\phi}(\bm{s})$ are the output action (torque) from the LQR and the hybrid policy $\phi(a|\bm{s})$ at given state $\bm{s}$, respectively. As seen in \cite{johannink2019residual,rana2020multiplicative}, using prior knowledge of the system (e.g., its model and conventional controller) can aid the RL network in operating within safer bounds as well as increase its sampling efficiency. Conversely, RL policies can assist conventional controller in adapting to various environmental changes by interacting with the world. The proposed \emph{Hybrid LMC} follows this outline but differs from previously explored residual RL frameworks - instead of a deterministic policy, a stochastic approach with distributional actions is utilized. We assume that a stochastic approach promotes the search - through randomly sampled behaviors - of the nearby action-space for more optimal torques. This ultimately result in better tracking of the desired states and in reduction of residual error, $\Delta \bm{q} = \bm{q}^{des} - \bm{q}$ created by unexpected disturbance, and environmental changes. The detailed procedure of \emph{Hybrid LMC} is decribed in Algorithm \ref{algo:blah}. Also we note that our strategy builds its action as the sum of a stochastic policy and the conventional feedback controller (i.e the LQR), unlike BCF where the action is only sampled from the hybrid policy \cite{rana2021bayesian}. 

\indent In order to take advantage of a stochastic approach and alleviate its drawback of large behavior variance, we use an ensemble technique that leverages multiple RL policy networks $\pi(a|\bm{s})$ in parallel \cite{chua2018deep}. The action of the $\phi(a|\bm{s})$ follows the composite Gaussian distribution $\phi(a|\bm{s}) \sim \mathcal{N}(\mu_{\phi},\,\sigma^{2}_{\phi})$ computed as follows:



\begin{equation}
\label{eqn_composite}
\mu_{\phi} = \dfrac{\mu_{\Pi}\sigma^{2}_{H}+\mu_{H}\sigma^{2}_{\Pi}}{\sigma^{2}_{\Pi}+\sigma^{2}_{H}} \;,\;
\sigma^{2}_{\phi}= \dfrac{\sigma^{2}_{H}\sigma^{2}_{\Pi}}{\sigma^{2}_{H}+\sigma^{2}_{\Pi}}
\end{equation}

\noindent where $\mu_{H}(\bm{s})$ denotes the mean of action from LQR and $\sigma^{2}_{H}$ is its variance. To acquire a distributional action from a conventional controller, we empirically assume the variance $\sigma^{2}_{H} (=0.4)$ for the LQR. The mean $\mu_{H}(\bm{s})$ is the same with an original action from LQR. We believe that the LQR, $H(a|\bm{s})$, can guide $\phi(a|\bm{s})$ in exploring more realistic torques during the early stages of training as the feedback controller is able to leverage prior information about the model and dynamics. 

As seen in ensemble techniques \cite{lakshminarayanan2017simple}, the mean $\mu_{\Pi}$ and variance $\sigma^{2}_{\Pi}$ of a uniformly weighted Gaussian mixture model $\Pi(a|\bm{s})$, Ensemble policy, are obtained by combining $M$ number of single RL policy $\pi(a|\bm{s})$:

\begin{equation}
\label{eqn_gmm1}
\mu_{\Pi}(\bm{s}) = M^{-1}\sum_{m=1}^{M}{\mu_{\pi_{m}}(\bm{s})} 
\end{equation}

\begin{equation}
\label{eqn_gmm2}
\sigma^{2}_{\Pi}(\bm{s})= M^{-1}\sum_{m=1}^{M}{(\sigma^{2}_{\pi_{m}}(\bm{s})+\mu^{2}_{\pi_{m}}(\bm{s}))}-\mu^{2}_{\Pi}(\bm{s})
\end{equation}

\noindent where $\mu_{\pi_m}(\bm{s})$ and $\sigma^{2}_{\pi_m}(\bm{s})$ denote the mean and variance of a single RL policy $\pi(a|\bm{s})$. Each RL policy $\pi(a|\bm{s})$ is trained with the use of the SAC algorithm \cite{haarnoja2018soft} that has achieved state-of-the-art (SOTA) performance in simulated robotic systems by addressing the continuous action problem. SAC was determined suitable here because it is a stochastic policy that chooses an action by sampling from a Gaussian distribution. This enables exploration of a larger state-space and action-space area.   


\begin{algorithm}[ht]
  \caption{Hybrid Leaning and Model-based Controller}\label{algo:blah}
  \begin{algorithmic}[1]
    \REQUIRE {Learned $M$ policies $(\pi_1(a_1|s_1),$ $\pi_2(a_2|s_2),$ $...$ $,\pi_M(a_M|s_M))$ 
    from SAC models and LQR controller $H(a|s) \sim \mathcal{N}(\mu_{H},\,\sigma^{2}_{H})$} 
    \ENSURE {Compensated torque $\tau_c$} 
    \STATE \textbf{for} $n=0,...$ epoch \textbf{do} 
    \begin{ALC@g}
    \STATE Select a single agent randomly among $M$ policies
    \STATE \textbf{for} $m=0,...,M$ multiple agents \textbf{do}
    \INDSTATE  Observe state $s_m$ and act an action $a_m$$\sim \pi_m(\cdot|s_m)$
    \STATE \textbf{end for}
    \STATE Compute a single univariate Gaussian distribution $\Pi(a|s) \sim \mathcal{N}(\mu_{\Pi},\,\sigma^{2}_{\Pi})$ (see Equation \ref{eqn_gmm1} and \ref{eqn_gmm2})
    \STATE Compute the composite Gaussian distribution $\phi(a|s) \sim \mathcal{N}(\mu_{\phi},\,\sigma^{2}_{\phi})$  (see Equation \ref{eqn_composite})
    
    \STATE $a = H(a|s) + \tanh{(\phi(a|s))}$
    
    \STATE Execute $a$ and Observe next state $s^{'}$, reward $r$, and done signal $d$
    \STATE Store $(s,a,r,s^{'},d)$ in replay buffer $\mathcal{D}$
    \STATE If $s'$ is terminal, reset environment state.
    \STATE \textbf{If} it's time to update \textbf{then}
    \begin{ALC@g}
    \STATE Compute targets and Update Q-function, policy, and target networks based on SAC algorithm \cite{haarnoja2018soft}
    \end{ALC@g}
    \textbf{end if}
    \end{ALC@g}
  \STATE \textbf{until} convergence
  \end{algorithmic}
\end{algorithm}

\indent The hybrid policy $\phi(a|\bm{s})$ samples the appropriate torques mainly affected by $\sigma^{2}_{\Pi}(\bm{s})$ and $\sigma^{2}_{H}(\bm{s})$. We assume that the variance of $\Pi(a|\bm{s})$ gradually decreases so $\phi(a|\bm{s})$ follows the ensemble policy $\Pi(a|\bm{s})$ more, and the LQR less as the policy is learned over time. This is motivated by epistemic uncertainty estimation techniques \cite{rana2021bayesian,lakshminarayanan2017simple}.  


\section{Experiment}

\subsection{Experimental Setup}
\label{experimetA}
\begin{figure}[t]
\begin{center}
\includegraphics[width=1\linewidth]{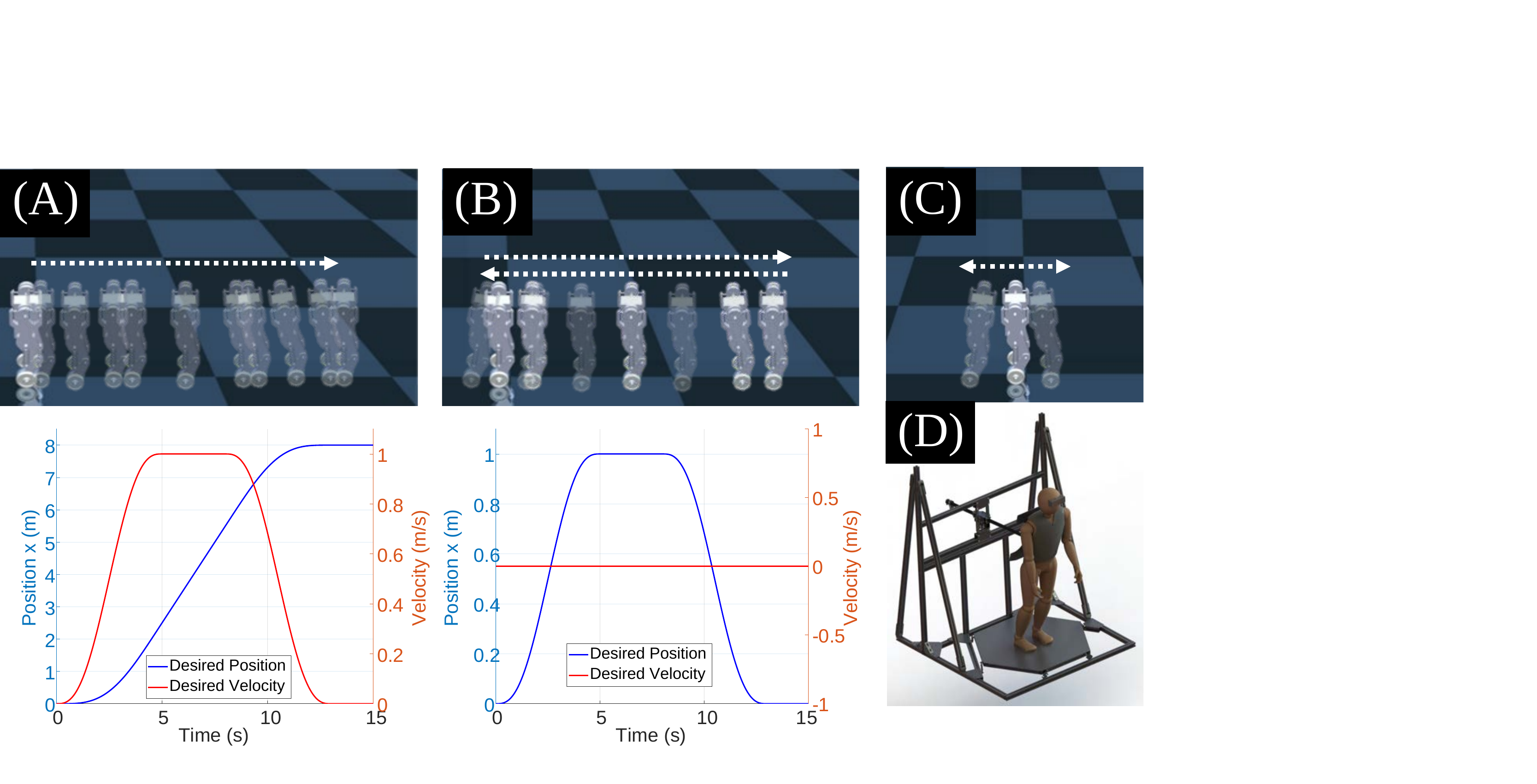}
\end{center}
   \caption{\textbf{Experiments in MuJoCo Simulation:} (A) Task2: a 5th-order velocity trajectory is given as an input $\dot{x}_w^{des}$ ($x_w^{des}=x_w^{des}+\dot{x}_w^{des}\Delta t$). (B) Task3: a 5th-order position trajectory is given as an input $x_w^{des}$ ($\dot{x}_w^{des}=0$). (C) Task1: Balancing task in a fixed point ($x_w^{des},\dot{x}_w^{des}=0$). (D) Human Machine Interface \cite{wang2021comparison} to provide a human data. $\bm{q}^{des}= [x_w^{des}, \dot{x}_w^{des}, \theta^{des},\dot{\theta}^{des}]^\top$ where both $\theta^{des}$ and $\dot{\theta}^{des}$ are $0$.} 
\label{figMujoco}
\end{figure}

\noindent \textbf{Simulation Setup:} All experiments for validating the \emph{Hybrid LMC} were conducted using MuJoCo \cite{todorov2012mujoco} simulation which is widely used to evaluate many learning-based methods. We modeled a wheeled humanoid robot, SATYRR using a Unified Robot Description Format (URDF) that has the same physical parameters (Table. \ref{t_specification}) as a real hardware platform. (Fig. \ref{figMujoco}) \\

\noindent \textbf{Experimental Variation:} We tested \emph{Hybrid LMC} on a diverse set of model parameters through changing of mass, friction, gear ratio, and CoM position values. The ranges of each parameter are described in Table. \ref{t_benchmark_pos}. \\

\noindent \textbf{Baselines:} \emph{Hybrid LMC} is compared to the following baselines:\\
1) LQR controller: LQR controller derived using a WIP model (a conventional feedback controller). \\
2) Model-free RL algorithms: SAC (stochastic approach) and Deep Deterministic Policy Gradient (DDPG) \cite{lillicrap2015continuous} (deterministic approach) have shown promising results in the continuous action space. \\
3) Residual RL: Residual RL framework \cite{johannink2019residual} consisting of the sum of a deterministic residual policy and a feedback controller. To benchmark this framework's perfromance for comparision, we tested DDPG with LQR (\emph{DDPG+LQR}) and SAC with LQR (\emph{SAC+LQR}) in our experiments.\\
4) Bayesian controller fusion (BCF): A hybrid control strategy combining a model-free RL and a conventional controller \cite{rana2021bayesian} that motivates a basic structure of \emph{Hybrid LMC}. 

\label{experimet}

\indent During experiments, we chose the best model from each method. All methods are trained with the same reward function, same state definition $s$, same LQR gains, and system parameters mentioned in section \ref{method}. Note that we only trained our model, \emph{Hybrid LMC},  using velocity profiles of 5th-order polynomials. Other methods (e.g. DDPG, SAC+LQR, etc.) required training on full reference trajectories. 

\noindent \textbf{Evaluation Metric:}
To compare the performance of \emph{Hybrid LMC} against baselines, we use root mean square error (RMSE) between  the errors of respective desired states and position of $x_w$, velocity $\dot{x}_w$, and pitch angle $\theta$. 

\subsection{Training Details}
\label{experimetB}
\noindent \textbf{Defining State-Action Space and Reward Function:} 
The state $\bm s$ at time $k$ is defined as $\bm{s}^k=\langle \bm{q}^k, \bm{\tau}^{k-1}, \bm{C}^k, \bm{\Theta}^k \rangle$ and each components are defined as follows: 
\begin{itemize}
    \item State-space vector: $\bm{q}^k$ =  $\langle x^k_w, \theta^k, \dot{x}^k_w, \dot{\theta}^k \rangle$
    \item Applied torque vector at the previous time step $t-1$: 
\indent $\bm{\tau}^{k-1} = \langle \tau^{k-1}_{LQR},\tau^{k-1}_{\phi}, \tau^{k-1}_c \rangle$
    \item The desired position and pitch angle:\\ 
\indent $\bm{C}^k = \langle C^k_x, C^k_{\theta} \rangle \in \bm{q}^{des}$ 
    \item History of position error and velocity:  
    $\bm{\Theta}^k = \langle \bm{\Theta}^k_{x_e},\bm{\Theta}^k_{\dot{x}} \rangle$
    $\bm{\Theta}^k_{x_e}= \langle \Delta x_w^{t-2},\Delta x_w^{t-1},\Delta x_w^{t} \rangle$, $\bm{\Theta}^k_{\dot{x}}= \langle \dot{x}^{t-2}_w, \dot{x}^{t-1}_w, \dot{x}^{t}_w \rangle$
\end{itemize}

\noindent where the symbol $\Delta$ indicates the error between the desired value and the actual value. The usage of $\bm{\tau}^{k-1}$ and  $\bm{\Theta}^k_{x_e}$ is motivated by previous works \cite{iscen2018policies,hwangbo2019learning}. The key to generating an end-to-end (state-to-torque) policy was found in the inclusion and usage of both $\bm{\tau}^{k-1}$ and $\bm{\Theta}^k_{x_e}$ within the residual RL framework.

The reward function was designed to track the robot's desired position $x^{des}_w$ and pitch $\theta$ to keep the robot stable. The resulting reward function $R$ at time $t$  is defined as follows : 

\begin{equation}
\label{eqn_reward1}
\begin{aligned}
R(s) = - K||\bm{e}^s_{t}||_2 +  \mathbf{1}{(|err(x_w)_t|<1)}*\mathbf{1}{(|err(\theta)_t|<.35)} \\ 
+ \mathbf{1}{(|err(x_w)_t| < |err'(x_w)_t|)}*\mathbf{1}{(|err(\theta)_t| < |err'(\theta)_t|)} 
\end{aligned}
\end{equation}

\noindent where $err(x)_t = x^{des}_w(t)-x_w(t)$ and $err(\theta)_t = 0-\theta(t)$. Indicating the change pattern of the error denotes $err'(y)_{t-1}=$ $y^{des}(t)-y(t-1)$. The scaling factor $K=[0.1, 0.1]$ and $\bm{e}^s_{t} = [err(x)_t, err(\theta)_t]^T$. 

\begin{figure*}[t]
     \centering
 	\begin{subfigure}{0.28\linewidth}
 		\includegraphics[width=\columnwidth]{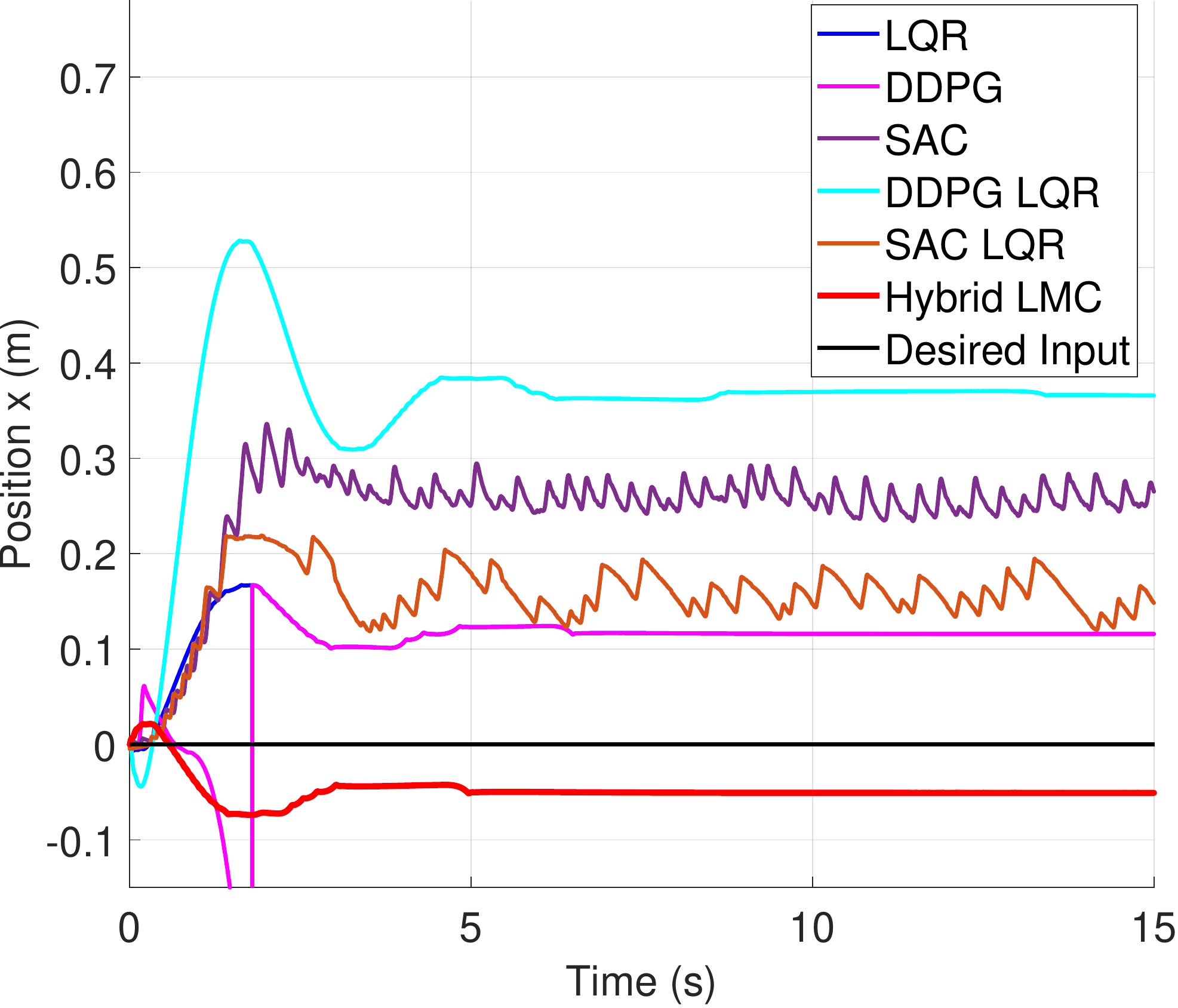}
 		\caption{}
 	\end{subfigure} 
 	\begin{subfigure}{0.28\linewidth}
 		\includegraphics[width=\columnwidth]{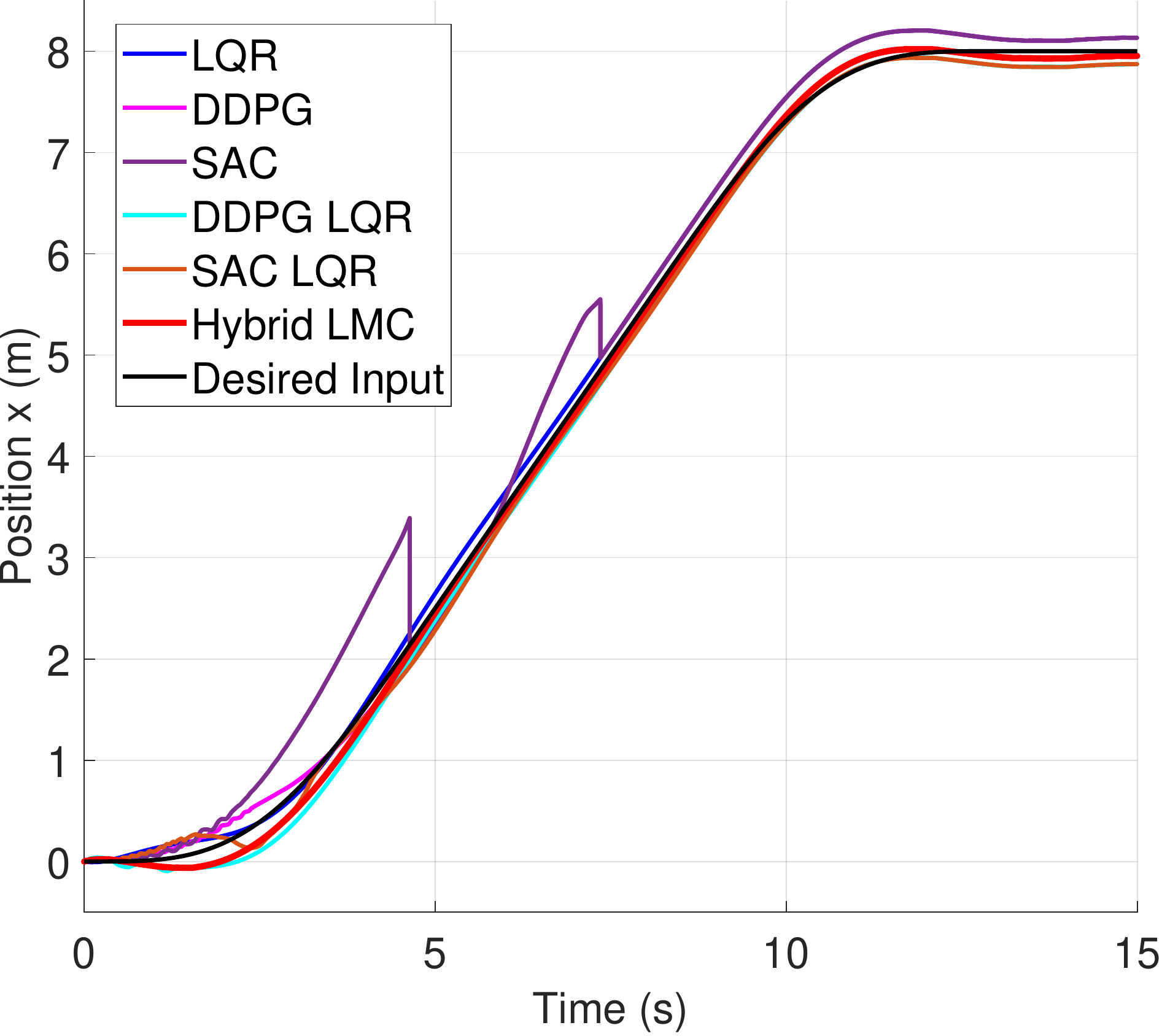}
 		\caption{}
 	\end{subfigure}
 	\begin{subfigure}{0.28\linewidth}
  		\includegraphics[width=\columnwidth]{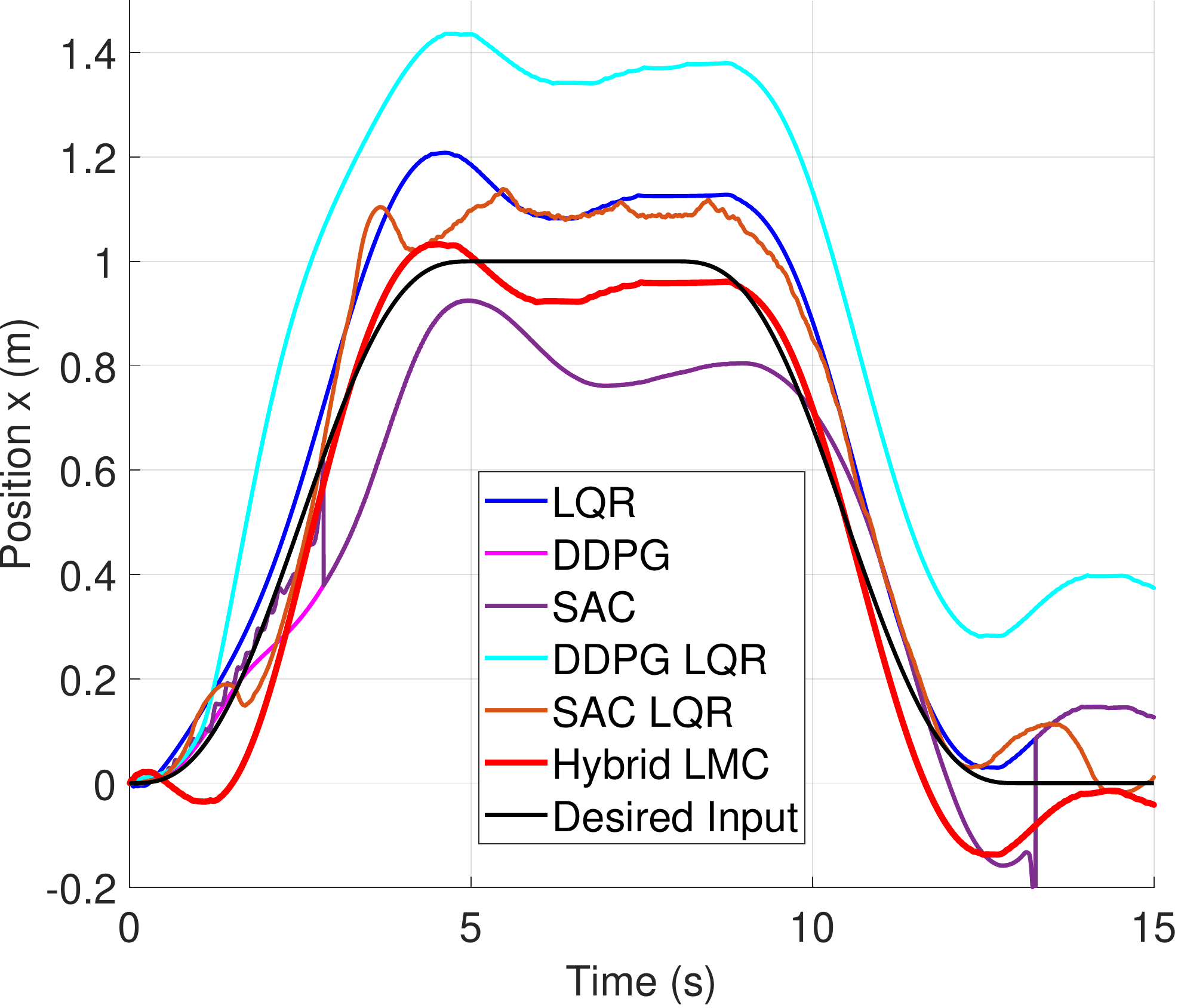}
 		\caption{}
 	\end{subfigure}

 	\begin{subfigure}{0.28\linewidth}
 		\includegraphics[width=\columnwidth]{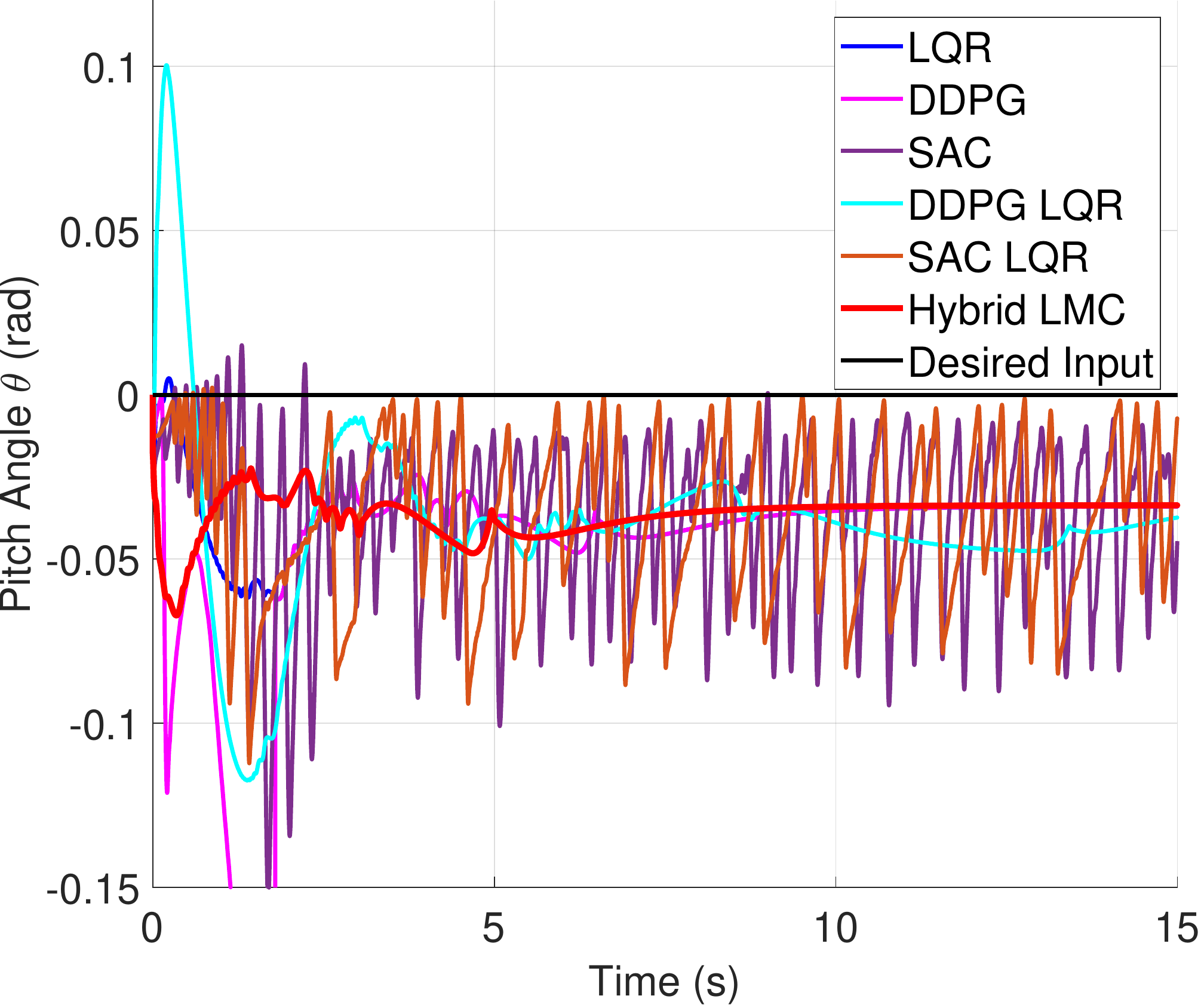}
 		\caption{}
 	\end{subfigure} 
 	\begin{subfigure}{0.28\linewidth}
 		\includegraphics[width=\columnwidth]{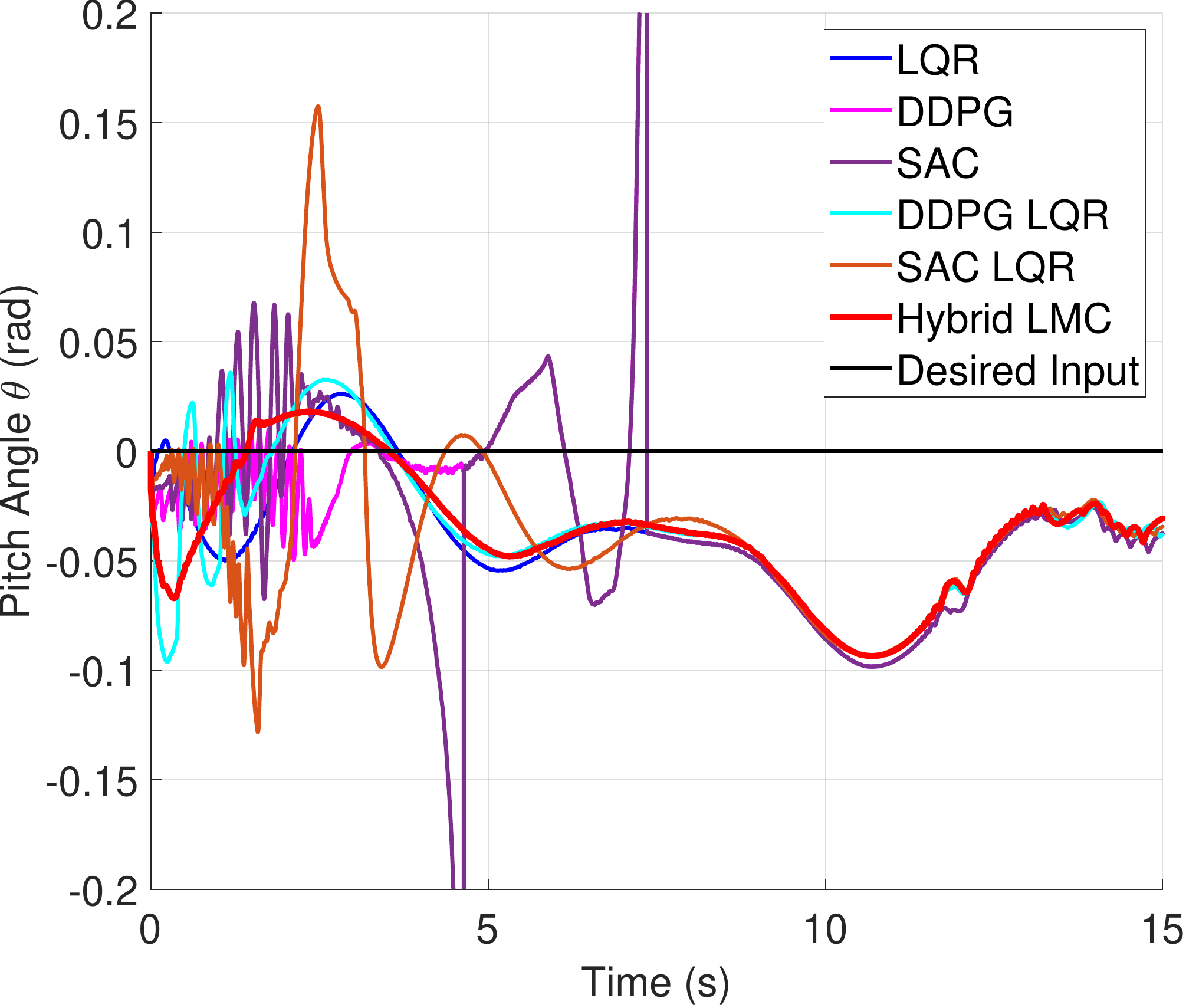}
 		\caption{}
 	\end{subfigure}
 	\begin{subfigure}{0.28\linewidth}
  		\includegraphics[width=\columnwidth]{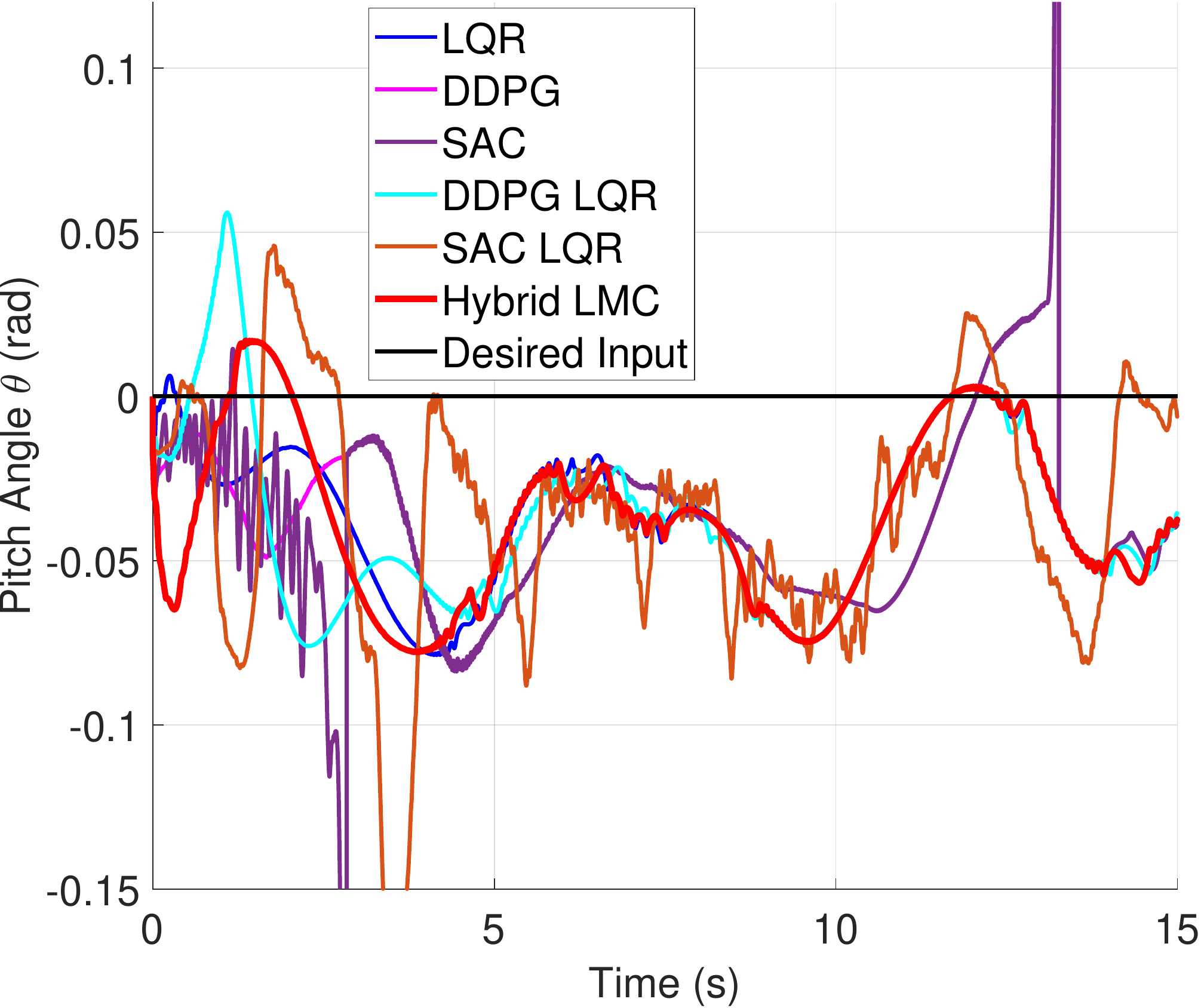}
 		\caption{}
 	\end{subfigure}

 	\caption{\textbf{Result of Locomotion Benchmark in Different Tasks}. Two figures in each column (e.g., (a) and (e)) indicate the position tracking and balancing performance of each task. Each column corresponds to task1, 2, and 3 from the leftmost. }
 	\label{benchmark_fig}
\end{figure*}

\begin{table*}[!htp]
\centering
\caption{\textbf{Locomotion Performance Benchmark.} Each table shows the experimental result performed in different tasks: (1) Balancing task (2) Tracking task given 5th-order velocity trajectory (3) Tracking task given 5th-order position trajectory. Normal case is set the same physical parameter with training environment. Case 2 and 3 are set with different physical parameters. The result value indicates the average of a total of 5 trials. The number in parentheses represents the probability of the operation time was performed when it failed. }
\label{t_benchmark_pos}
\setlength{\tabcolsep}{5.1pt}
\begin{tabular}{|c|c|ccccccc|}
\hline
\multirow{2}{*}{\textbf{\begin{tabular}[c]{@{}c@{}}Model Settings \\ (Mass, Gear ratio, Friction, CoM)\end{tabular}}} & \multirow{2}{*}{\textbf{\begin{tabular}[c]{@{}c@{}}Metric \\ (RMSE)\end{tabular}}} & \multicolumn{7}{c|}{\textbf{Method}}                                                                                                                                                                                                                \\ \cline{3-9} 
                                                                                                                      &                                                                                    & \multicolumn{1}{c|}{\textbf{LQR}} & \multicolumn{1}{c|}{\textbf{DDPG}} & \multicolumn{1}{c|}{\textbf{SAC}} & \multicolumn{1}{c|}{\textbf{DDPG+LQR}} & \multicolumn{1}{c|}{\textbf{SAC+LQR}} & \multicolumn{1}{l|}{\textbf{BCF}} & \textbf{Hybrid LMC} \\ \hline
\multirow{3}{*}{\textbf{\begin{tabular}[c]{@{}c@{}}Case 1 (Normal)\\ (4.05, 1 ,1, 0)\end{tabular}}}                   & Position (m)                                                              & \multicolumn{1}{c|}{0.128}        & \multicolumn{1}{c|}{\textbf{0.025}}         & \multicolumn{1}{c|}{0.271}        & \multicolumn{1}{c|}{0.377}             & \multicolumn{1}{c|}{0.164}            & \multicolumn{1}{c|}{F}            & 0.038   \\ \cline{2-9} 
                                                                                                                      & Velocity (m/s)                                                            & \multicolumn{1}{c|}{0.06}         & \multicolumn{1}{c|}{0.283}         & \multicolumn{1}{c|}{0.279}        & \multicolumn{1}{c|}{0.159}             & \multicolumn{1}{c|}{0.209}            & \multicolumn{1}{c|}{F}            & \textbf{0.043}    \\ \cline{2-9} 
                                                                                                                      & Pitch angle (rad)                                                         & \multicolumn{1}{c|}{0.042}        & \multicolumn{1}{c|}{0.051}         & \multicolumn{1}{c|}{0.048}        & \multicolumn{1}{c|}{0.048}             & \multicolumn{1}{c|}{0.049}            & \multicolumn{1}{c|}{F}            & \textbf{0.041}    \\ \hline
\multirow{3}{*}{\textbf{\begin{tabular}[c]{@{}c@{}}Case 2\\ (8.05, 1.3 ,1.3, 0.12)\end{tabular}}}                     & Position (m)                                                             & \multicolumn{1}{c|}{0.116}        & \multicolumn{1}{c|}{0.126 (0.12)}  & \multicolumn{1}{c|}{0.252}        & \multicolumn{1}{c|}{0.367}             & \multicolumn{1}{c|}{0.156}            & \multicolumn{1}{c|}{F}            & \textbf{0.05}     \\ \cline{2-9} 
                                                                                                                      & Velocity (m/s)                                                            & \multicolumn{1}{c|}{0.05}         & \multicolumn{1}{c|}{0.456 (0.12)}  & \multicolumn{1}{c|}{0.287}        & \multicolumn{1}{c|}{0.156}             & \multicolumn{1}{c|}{0.209}            & \multicolumn{1}{c|}{F}            & \textbf{0.046}    \\ \cline{2-9} 
                                                                                                                      & Pitch angle (rad)                                                         & \multicolumn{1}{c|}{0.038}        & \multicolumn{1}{c|}{0.181 (0.12)}  & \multicolumn{1}{c|}{0.045}        & \multicolumn{1}{c|}{0.047}             & \multicolumn{1}{c|}{0.044}            & \multicolumn{1}{c|}{F}            & \textbf{0.036}    \\ \hline
\multirow{3}{*}{\textbf{\begin{tabular}[c]{@{}c@{}}Case 3\\ (14.05, 0.9 ,1.1, -0.12)\end{tabular}}}                   & Position (m)                                                              & \multicolumn{1}{c|}{0.116}        & \multicolumn{1}{c|}{0.083 (0.12)}  & \multicolumn{1}{c|}{0.228}        & \multicolumn{1}{c|}{0.368}             & \multicolumn{1}{c|}{0.148}            & \multicolumn{1}{c|}{F}            & \textbf{0.053}    \\ \cline{2-9} 
                                                                                                                      & Velocity (m/s)                                                            & \multicolumn{1}{c|}{0.081}        & \multicolumn{1}{c|}{0.323 (0.12)}  & \multicolumn{1}{c|}{0.289}        & \multicolumn{1}{c|}{0.177}             & \multicolumn{1}{c|}{0.205}            & \multicolumn{1}{c|}{F}            & \textbf{0.074}    \\ \cline{2-9} 
                                                                                                                      & Pitch angle (rad)                                                         & \multicolumn{1}{c|}{0.039}        & \multicolumn{1}{c|}{0.146 (0.12)}  & \multicolumn{1}{c|}{0.063}        & \multicolumn{1}{c|}{0.050}             & \multicolumn{1}{c|}{0.047}            & \multicolumn{1}{c|}{F}            & \textbf{0.038}    \\ \hline
\end{tabular}
\vspace{0.2cm}

\setlength{\tabcolsep}{4pt}
\begin{tabular}{|c|c|ccccccc|}
\hline
\multirow{2}{*}{\textbf{\begin{tabular}[c]{@{}c@{}}Model Settings \\ (Mass, Gear ratio, Friction, CoM)\end{tabular}}} & \multirow{2}{*}{\textbf{\begin{tabular}[c]{@{}c@{}}Metric \\ (RMSE)\end{tabular}}} & \multicolumn{7}{c|}{\textbf{Method}}                                                                                                                                                                                                                  \\ \cline{3-9} 
                                                                                                                      &                                                                                    & \multicolumn{1}{c|}{\textbf{LQR}}   & \multicolumn{1}{c|}{\textbf{DDPG}} & \multicolumn{1}{c|}{\textbf{SAC}} & \multicolumn{1}{c|}{\textbf{DDPG+LQR}} & \multicolumn{1}{c|}{\textbf{SAC+LQR}} & \multicolumn{1}{l|}{\textbf{BCF}} & \textbf{Hybrid LMC} \\ \hline
\multirow{3}{*}{\textbf{\begin{tabular}[c]{@{}c@{}}Case 1 (Normal)\\ (4.05, 1 ,1, 0)\end{tabular}}}                   & Position (m)                                                                       & \multicolumn{1}{c|}{0.147}          & \multicolumn{1}{c|}{0.269 (0.47)}  & \multicolumn{1}{c|}{0.482 (0.26)} & \multicolumn{1}{c|}{0.137}             & \multicolumn{1}{c|}{0.116}            & \multicolumn{1}{c|}{F}            & \textbf{0.083}    \\ \cline{2-9} 
                                                                                                                      & Velocity (m/s)                                                                     & \multicolumn{1}{c|}{0.084}          & \multicolumn{1}{c|}{0.313 (0.47)}  & \multicolumn{1}{c|}{0.370 (0.26)} & \multicolumn{1}{c|}{0.109}             & \multicolumn{1}{c|}{0.184}            & \multicolumn{1}{c|}{F}            & \textbf{0.074}    \\ \cline{2-9} 
                                                                                                                      & Pitch angle (rad)                                                                  & \multicolumn{1}{c|}{\textbf{0.049}} & \multicolumn{1}{c|}{0.054 (0.47)}  & \multicolumn{1}{c|}{0.027 (0.26)} & \multicolumn{1}{c|}{0.051}             & \multicolumn{1}{c|}{0.059}            & \multicolumn{1}{c|}{F}            & \textbf{0.049}    \\ \hline
\multirow{3}{*}{\textbf{\begin{tabular}[c]{@{}c@{}}Case 2\\ (8.05, 1.3 ,1.3, 0.12)\end{tabular}}}                     & Position (m)                                                                       & \multicolumn{1}{c|}{0.136}          & \multicolumn{1}{c|}{0.285 (0.53)}  & \multicolumn{1}{c|}{0.583 (0.3)}  & \multicolumn{1}{c|}{0.146}             & \multicolumn{1}{c|}{0.122}            & \multicolumn{1}{c|}{F}            & \textbf{0.087}    \\ \cline{2-9} 
                                                                                                                      & Velocity (m/s)                                                                     & \multicolumn{1}{c|}{0.081}          & \multicolumn{1}{c|}{0.380 (0.53)}  & \multicolumn{1}{c|}{0.394 (0.3)}  & \multicolumn{1}{c|}{0.11}              & \multicolumn{1}{c|}{0.181}            & \multicolumn{1}{c|}{F}            & \textbf{0.074}    \\ \cline{2-9} 
                                                                                                                      & Pitch angle (rad)                                                                  & \multicolumn{1}{c|}{\textbf{0.046}} & \multicolumn{1}{c|}{0.048 (0.53)}  & \multicolumn{1}{c|}{0.070 (0.3)}  & \multicolumn{1}{c|}{0.048}             & \multicolumn{1}{c|}{0.056}            & \multicolumn{1}{c|}{F}            & \textbf{0.046}    \\ \hline
\multirow{3}{*}{\textbf{\begin{tabular}[c]{@{}c@{}}Case 3\\ (14.05, 0.9 ,1.1, -0.12)\end{tabular}}}                   & Position (m)                                                                       & \multicolumn{1}{c|}{0.129}          & \multicolumn{1}{c|}{0.265 (0.51)}  & \multicolumn{1}{c|}{0.05 (0.11)}  & \multicolumn{1}{c|}{0.14}              & \multicolumn{1}{c|}{0.122}            & \multicolumn{1}{c|}{F}            & \textbf{0.077}    \\ \cline{2-9} 
                                                                                                                      & Velocity (m/s)                                                                     & \multicolumn{1}{c|}{0.080}          & \multicolumn{1}{c|}{0.349 (0.51)}  & \multicolumn{1}{c|}{0.312 (0.11)} & \multicolumn{1}{c|}{0.109}             & \multicolumn{1}{c|}{0.188}            & \multicolumn{1}{c|}{F}            & \textbf{0.073}    \\ \cline{2-9} 
                                                                                                                      & Pitch angle (rad)                                                                  & \multicolumn{1}{c|}{\textbf{0.045}} & \multicolumn{1}{c|}{0.048 (0.51)}  & \multicolumn{1}{c|}{0.092 (0.11)} & \multicolumn{1}{c|}{0.048}             & \multicolumn{1}{c|}{0.057}            & \multicolumn{1}{c|}{F}            & 0.046             \\ \hline
\end{tabular}

\vspace{0.2cm}
\setlength{\tabcolsep}{4pt}
\begin{tabular}{|c|c|ccccccc|}
\hline
\multirow{2}{*}{\textbf{\begin{tabular}[c]{@{}c@{}}Model Settings \\ (Mass, Gear ratio, Friction, CoM)\end{tabular}}} & \multirow{2}{*}{\textbf{\begin{tabular}[c]{@{}c@{}}Metric \\ (RMSE)\end{tabular}}} & \multicolumn{7}{c|}{\textbf{Method}}                                                                                                                                                                                                                    \\ \cline{3-9} 
                                                                                                                      &                                                                                    & \multicolumn{1}{c|}{\textbf{LQR}}   & \multicolumn{1}{c|}{\textbf{DDPG}} & \multicolumn{1}{c|}{\textbf{SAC}} & \multicolumn{1}{c|}{\textbf{DDPG+LQR}} & \multicolumn{1}{c|}{\textbf{SAC+LQR}} & \multicolumn{1}{l|}{\textbf{BCF}} & \textbf{Hybrid LMC} \\ \hline
\multirow{3}{*}{\textbf{\begin{tabular}[c]{@{}c@{}}Case 1 (Normal)\\ (4.05, 1 ,1, 0)\end{tabular}}}                   & Position (m)                                                                       & \multicolumn{1}{c|}{0.140}          & \multicolumn{1}{c|}{0.144 (0.88)}  & \multicolumn{1}{c|}{0.122 (0.3)}  & \multicolumn{1}{c|}{0.374}             & \multicolumn{1}{c|}{0.110}            & \multicolumn{1}{c|}{F}            & \textbf{0.060}      \\ \cline{2-9} 
                                                                                                                      & Velocity (m/s)                                                                     & \multicolumn{1}{c|}{\textbf{0.082}} & \multicolumn{1}{c|}{0.202 (0.88)}  & \multicolumn{1}{c|}{0.336 (0.3)}  & \multicolumn{1}{c|}{0.120}             & \multicolumn{1}{c|}{0.175}            & \multicolumn{1}{c|}{F}            & 0.087               \\ \cline{2-9} 
                                                                                                                      & Pitch angle (rad)                                                                  & \multicolumn{1}{c|}{\textbf{0.047}} & \multicolumn{1}{c|}{0.051 (0.88)}  & \multicolumn{1}{c|}{0.091 (0.3)}  & \multicolumn{1}{c|}{0.049}             & \multicolumn{1}{c|}{0.057}            & \multicolumn{1}{c|}{F}            & \textbf{0.047}      \\ \hline
\multirow{3}{*}{\textbf{\begin{tabular}[c]{@{}c@{}}Case 2\\ (8.05, 1.3 ,1.3, 0.12)\end{tabular}}}                     & Position (m)                                                                       & \multicolumn{1}{c|}{0.128}          & \multicolumn{1}{c|}{0.152 (0.88)}  & \multicolumn{1}{c|}{0.04 (0.19)}  & \multicolumn{1}{c|}{0.362}             & \multicolumn{1}{c|}{0.100}            & \multicolumn{1}{c|}{F}            & \textbf{0.058}      \\ \cline{2-9} 
                                                                                                                      & Velocity (m/s)                                                                     & \multicolumn{1}{c|}{\textbf{0.080}} & \multicolumn{1}{c|}{0.209 (0.88)}  & \multicolumn{1}{c|}{0.326 (0.19)} & \multicolumn{1}{c|}{0.120}             & \multicolumn{1}{c|}{0.179}            & \multicolumn{1}{c|}{F}            & 0.087               \\ \cline{2-9} 
                                                                                                                      & Pitch angle (rad)                                                                  & \multicolumn{1}{c|}{\textbf{0.043}} & \multicolumn{1}{c|}{0.048 (0.88)}  & \multicolumn{1}{c|}{0.07 (0.19)}  & \multicolumn{1}{c|}{0.046}             & \multicolumn{1}{c|}{0.054}            & \multicolumn{1}{c|}{F}            & \textbf{0.043}      \\ \hline
\multirow{3}{*}{\textbf{\begin{tabular}[c]{@{}c@{}}Case 3\\ (14.05, 0.9 ,1.1, -0.12)\end{tabular}}}                   & Position (m)                                                                       & \multicolumn{1}{c|}{0.122}          & \multicolumn{1}{c|}{0.154 (0.87)}  & \multicolumn{1}{c|}{0.041 (0.2)}  & \multicolumn{1}{c|}{0.359}             & \multicolumn{1}{c|}{0.101}            & \multicolumn{1}{c|}{F}            & \textbf{0.046}      \\ \cline{2-9} 
                                                                                                                      & Velocity (m/s)                                                                     & \multicolumn{1}{c|}{\textbf{0.082}} & \multicolumn{1}{c|}{0.205 (0.87)}  & \multicolumn{1}{c|}{0.213 (0.2)}  & \multicolumn{1}{c|}{0.132}             & \multicolumn{1}{c|}{0.243}            & \multicolumn{1}{c|}{F}            & \textbf{0.082}      \\ \cline{2-9} 
                                                                                                                      & Pitch angle (rad)                                                                  & \multicolumn{1}{c|}{0.044}          & \multicolumn{1}{c|}{0.047 (0.87)}  & \multicolumn{1}{c|}{0.084 (0.2)}  & \multicolumn{1}{c|}{0.047}             & \multicolumn{1}{c|}{0.067}            & \multicolumn{1}{c|}{F}            & \textbf{0.043}      \\ \hline
\end{tabular}

\end{table*}

\noindent \textbf{Learning the Ensemble Deep Reinforcement Learning:} In this work, we use 10 single SAC policy $\pi(a|\bm{s})$ ($M=10$). A single RL policy $\pi(a|\bm{s})$ is a 3-layer multi-layer perceptron (MLP), with input $\bm{s} \in \R^{15}$, and output $u \in \R^1$ that represents wheel torque for stabilization. We trained for total of 5,00 epochs with each epoch consisting of a maximum of 4,000 steps. The policy was updated at every 1,000 steps. The size of our replay buffer was $1e^6$. The learning rate was set to $1e^{-3}$. The discount factor was set to 0.99. We chose $\rho$, for updating our target network value, to be 0.995. The entropy regularization coefficient value is chosen using an auto temperature adjustment \cite{pertsch2020accelerating}. The batch size is 64 and the Adam optimizer was utilized for all our implementations. The structure of SAC was implemented by referring the OPENAI open source \cite{brockman2016openai}. A single desktop with 1 GPU (RTX3060ti) was used for training RL algorithms. Training takes roughly half day on the desktop machine.

\section{Result and Discussion}
\label{result}

In this section we conducted three tests to gauge the performance of the proposed \emph{Hybrid LMC}: A) A performance benchmark. B) \emph{Hybrid LMC} with human control data test. C) An ablation study highlighting key design parameters of \emph{Hybrid LMC}.

In section \ref{PerfBenchmark} we fix our model parameters and reward function and compare against different controllers ranging from the standard LQR, model-free RLs, a residual RLs, to \emph{Hybrid LMC}. We showcase our results and discuss why we believe \emph{Hybrid LMC} demonstrates improved performance. In section \ref{HMICaseStudy}, we highlight the performance of \emph{Hybrid LMC} in the absence of hand-designed preconstructed trajectories, using human HMI commands for reference instead. Finally, in section \ref{AblationStudy} we discuss an ablation study with varied design parameters, specifically using data history and previous time step torque values as an input of the network.


\subsection{Performance Benchmark} \label{PerfBenchmark}
We have summarized the performance of \emph{Hybrid LMC} in Table \ref{t_benchmark_pos} and described in Fig \ref{benchmark_fig}. Overall, \emph{Hybrid LMC} achieves the highest performance (RMSE values less than 0.1) in all test cases. The proposed \emph{Hybrid LMC} shows an average performance increase of 48\% for position tracking compared to just an LQR. There was no significant improvement in tracking of the other states, $\dot{x}_w$ and $\theta$. Interestingly, \emph{Hybrid LMC} showed better overall performance for the 3 different tracking tasks, specified in \ref{t_benchmark_pos}, despite being trained using only desired velocity trajectories and their integrated terms, as described in Fig. \ref{figMujoco}. Basic residual RL was not able to enhance performance and even performed worse than just the LQR in certain situations. Lastly, the policy with only BCF failed to generate the appropriate torques for completing the locomotion tasks and stabilizing SATYRR.

\indent In our studies we consistently found that the LQR controller had larger position steady state error than steady state pitch error. \emph{Hybrid LMC} is attempting to reduce the residual error by rewarding smaller deviations from the desired. Hence, we hypothesize that our policy effectively learned to complement the LQR and help reduce the largest source of error found in position tracking. As the other errors in pitch and velocity were already small, we noticed marginal improvement. 

\indent For the balancing task, model-free DDPG had the best performance for Task 1, Case 1. In other cases, DDPG failed to successfully track the desired trajectories. Also, the performance of \emph{DDPG+LQR} significantly decreased with changes in the model parameters (mass, friction, etc.). The stochastic approach, \emph{SAC+LQR}, showed similar successful performance regardless of the model parameter changes. This suggests that a stochastic approach might be more advantageous in building a versatile policy against model changes for locomotion. We believe that stochastic policies promote exploration of the action space through random sampling of the output action distribution, which in turn enables discovery of potentially more ideal actions. However, methods using a single stochastic policy, i.e. SAC, are prone to having large output variance. Consequently, this degrades control performance and consistency. The performance difference between \emph{Hybrid LMC} and \emph{SAC+LQR} suggests that a deep ensemble RL is efficient in handling the high-variance issues highlighted above (see Fig. \ref{benchmark_fig}). Finally, from our comparisons of model-free RL we also see that these policies do not effectively learn end-to-end, state-to-torque values for the wheeled robot locomotion problem. These initial findings highlight the need for hybrid polices such as \emph{Hybrid LMC}.

\begin{figure}[t]
    \centering
 	\begin{subfigure}{0.32\linewidth}
 		\includegraphics[width=\columnwidth]{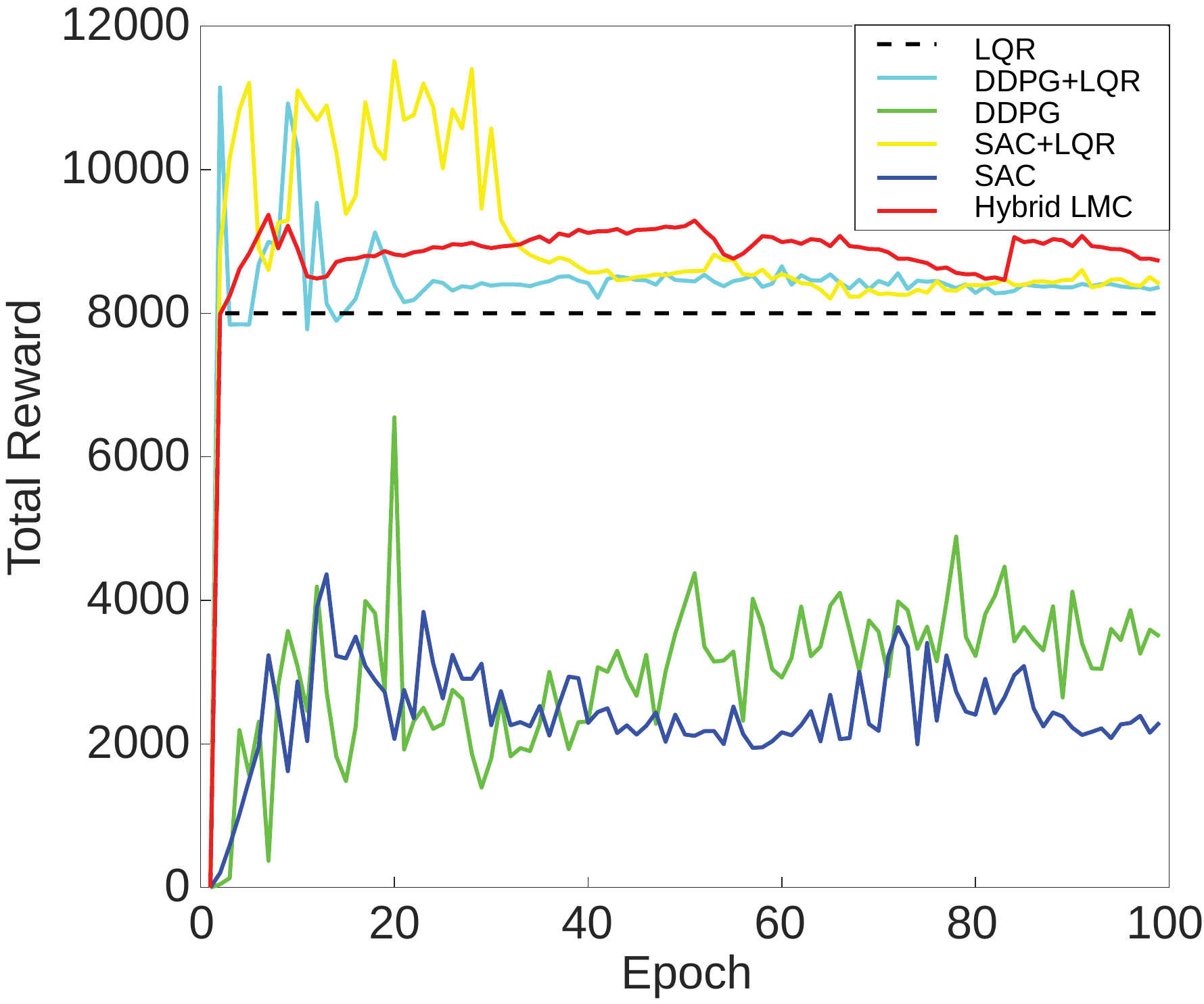}
 		\caption{}
 	\end{subfigure} 
 	\begin{subfigure}{0.32\linewidth}
 		\includegraphics[width=\columnwidth]{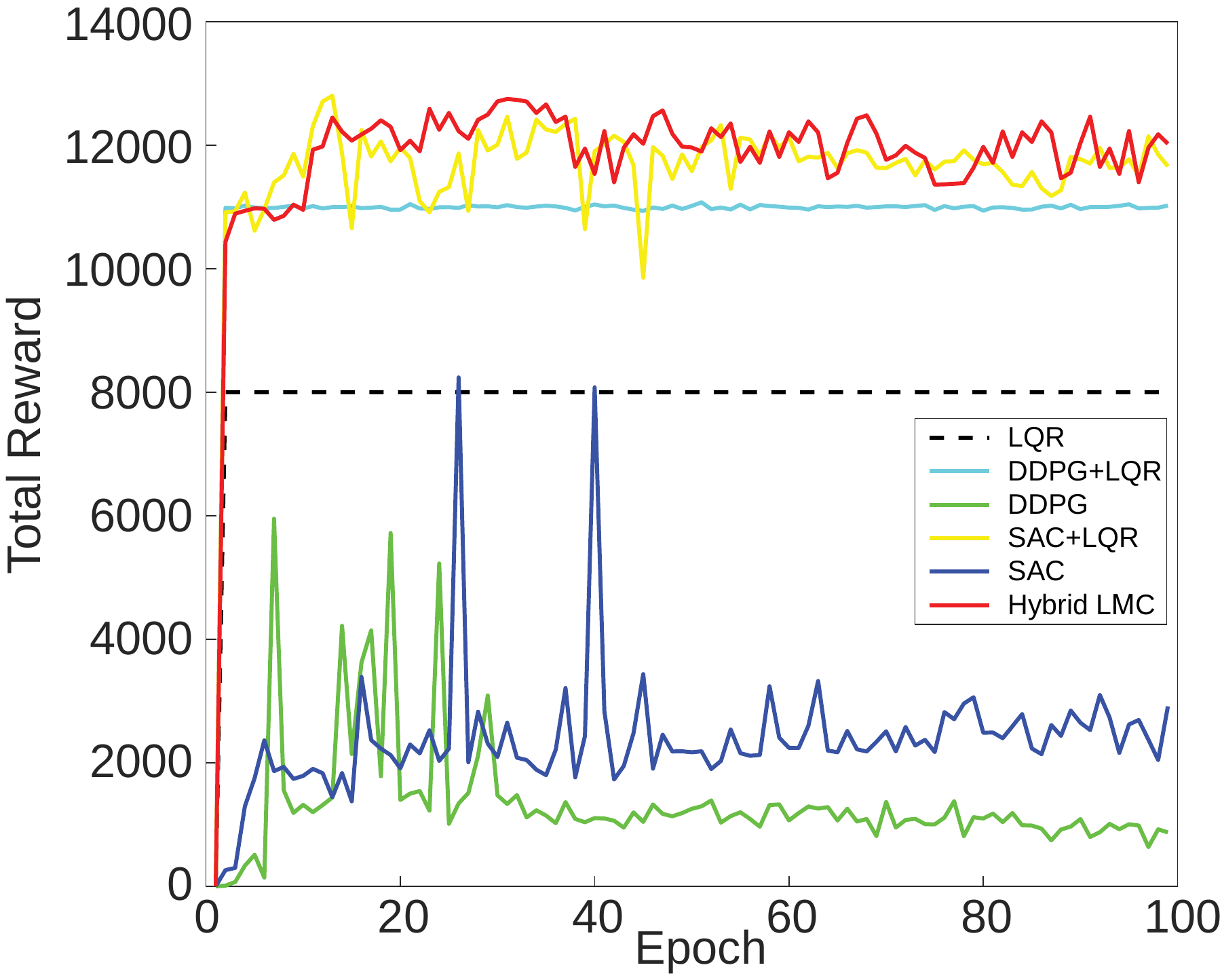}
 		\caption{}
    \end{subfigure}
    \begin{subfigure}{0.32\linewidth}
 		\includegraphics[width=\columnwidth]{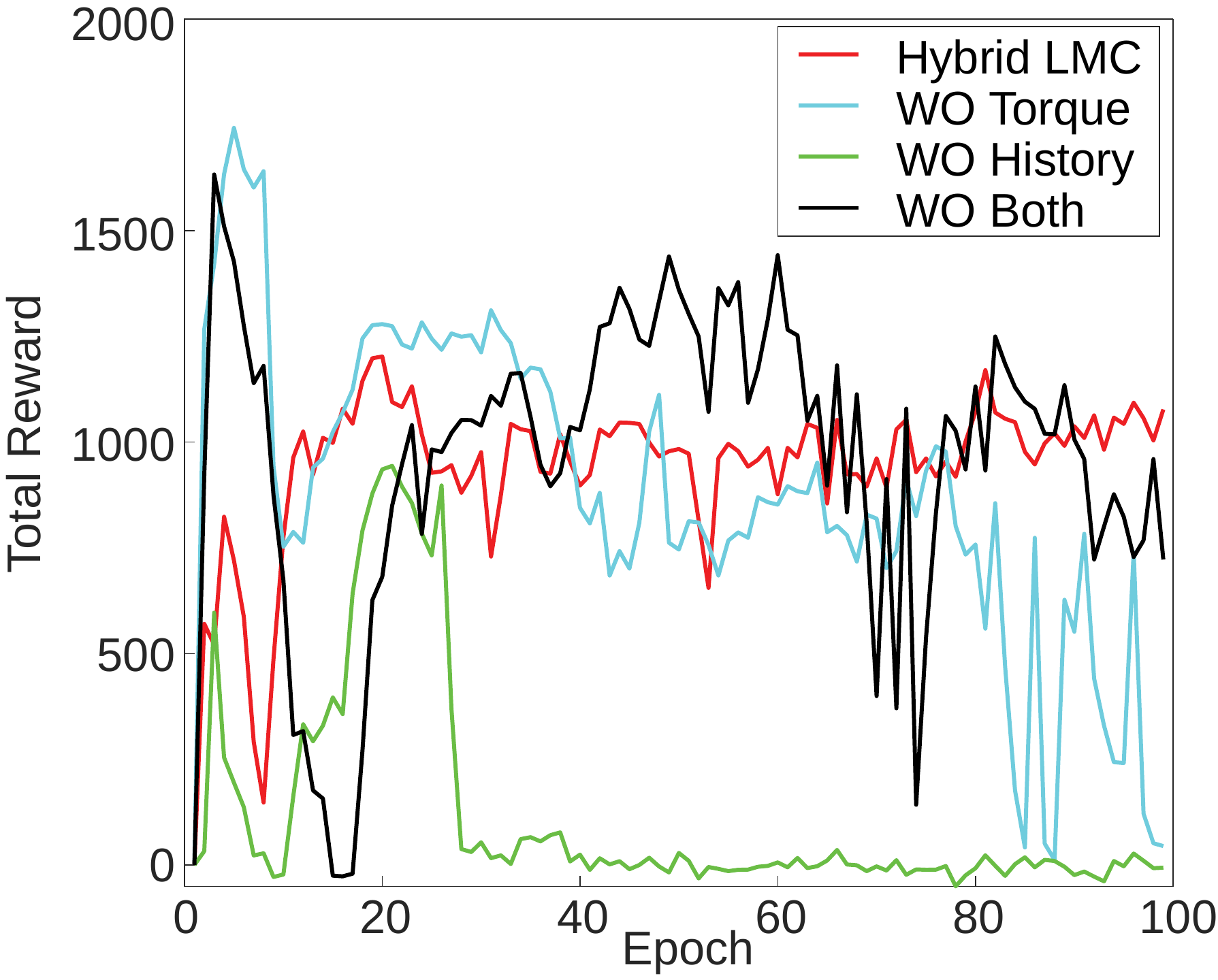}
 		\caption{}
    \end{subfigure}

\caption{\textbf{Learning Curves of the Total Reward.} (a) Total test reward graph given a 5th-order velocity trajectory. (b) Total test reward graph given command to stand upright in place. (c) Comparison total reward value for ablation study.}
\label{reward_fig}
\end{figure}

\begin{figure}[t]
\begin{center}
\includegraphics[width=1\linewidth]{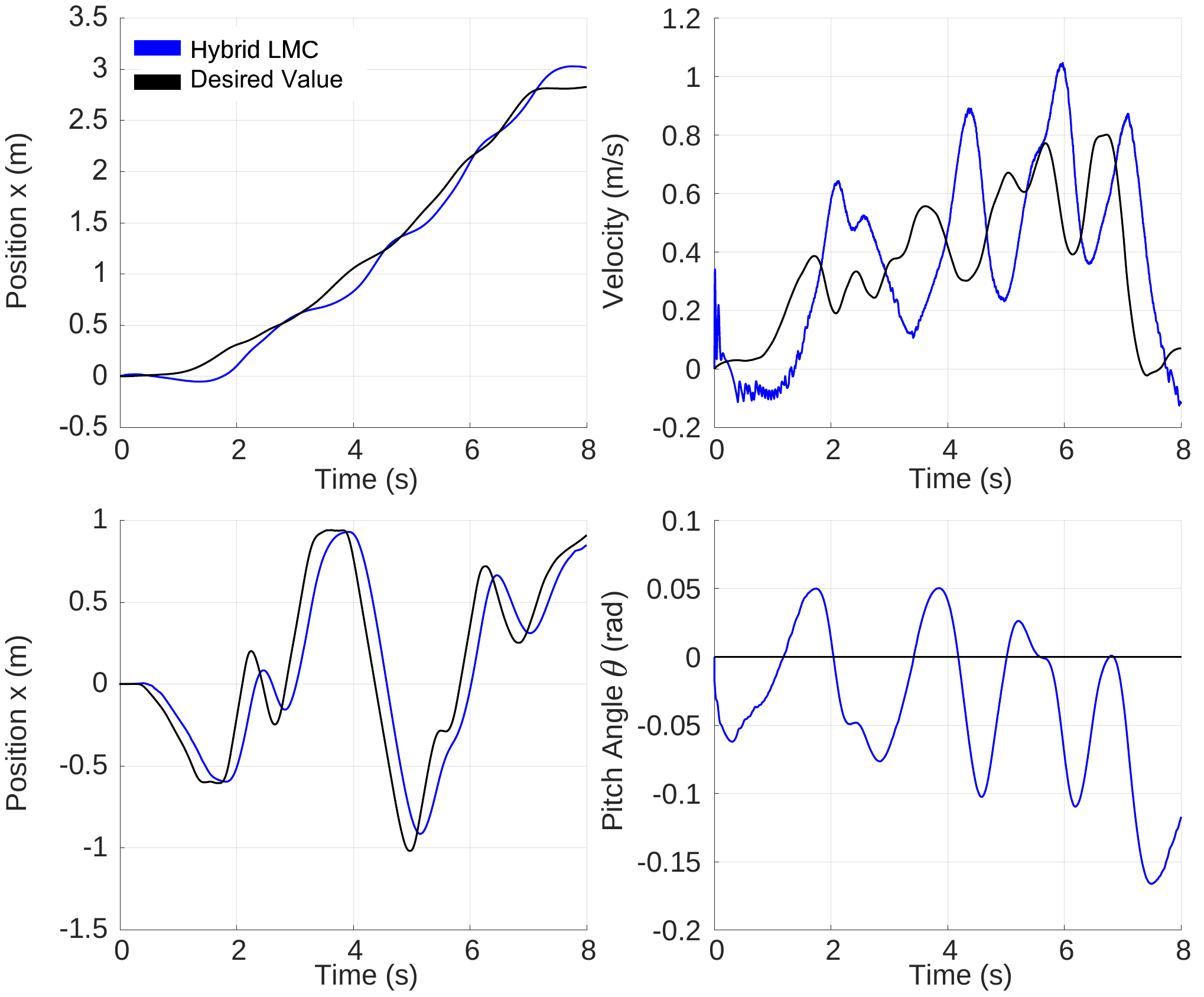}
\end{center}
   \caption{\textbf{Results of the Performance of Hybrid LMC using a Human Data.} RMSE of each state is 0.124, 0.273, 0.064, 0.2419, corresponding to position, velocity, pitch angle, and yaw position.}
\label{ablation_study_fig}
\end{figure}



\begin{table}[t]
\centering
\caption{\textbf{Ablation Study for Hybrid LMC:} Performance result without a key component.}
\label{t_ablation_state}
\setlength{\tabcolsep}{0.75pt}
\begin{tabular}{|c|c|ccc|}
\hline
\multirow{2}{*}{\textbf{}}      & \multirow{2}{*}{\textbf{Hybrid LMC}} & \multicolumn{3}{c|}{\textbf{Metric (RMSE)}}                                                                            \\ \cline{3-5} 
                                &                                      & \multicolumn{1}{c|}{\textbf{Position (m)}} & \multicolumn{1}{c|}{\textbf{Velocity (m/s)}} & \textbf{Pitch angle (rad)} \\ \hline
\multirow{4}{*}{\textbf{Task1}} & W/O history data                     & \multicolumn{1}{c|}{0.132}                 & \multicolumn{1}{c|}{0.074}                   & 0.042                      \\ \cline{2-5} 
                                & W/O LQR torque                       & \multicolumn{1}{c|}{0.108}                 & \multicolumn{1}{c|}{0.067}                   & \textbf{0.041}             \\ \cline{2-5} 
                                & W/O Both                             & \multicolumn{1}{c|}{0.149}                 & \multicolumn{1}{c|}{0.095}                   & 0.042                      \\ \cline{2-5} 
                                & W Both                               & \multicolumn{1}{c|}{\textbf{0.032}}        & \multicolumn{1}{c|}{\textbf{0.048}}          & \textbf{0.041}             \\ \hline
\multirow{4}{*}{\textbf{Task2}} & W/O history data                     & \multicolumn{1}{c|}{0.134}                 & \multicolumn{1}{c|}{0.09}                    & \textbf{0.049}             \\ \cline{2-5} 
                                & W/O LQR torque                       & \multicolumn{1}{c|}{0.146}                 & \multicolumn{1}{c|}{0.09}                    & \textbf{0.049}             \\ \cline{2-5} 
                                & W/O Both                             & \multicolumn{1}{c|}{0.166}                 & \multicolumn{1}{c|}{0.113}                   & 0.05                       \\ \cline{2-5} 
                                & W Both                               & \multicolumn{1}{c|}{\textbf{0.079}}        & \multicolumn{1}{c|}{\textbf{0.076}}          & \textbf{0.049}             \\ \hline
\end{tabular}
\end{table}

\indent All learning curves are shown in Fig. \ref{reward_fig}. We trained all models until they converged and the total reward no longer increased. The methods based on residual RL (e.g. \emph{SAC+LQR} and \emph{Hybrid LMC}) showed faster convergence than their model-free counterparts (SAC, DPPG). These hybrid controllers converged in approximately 20 epochs. Compared to \emph{Hybrid LMC}, we found that \emph{SAC+LQR} achieved a larger reward value but also had larger variance here. This high variance can result in undesired noisy output actions as shown in Fig. \ref{benchmark_fig}.

\emph{Hybrid LMC} consistently showed improved performance in reducing residual error in simulation, but application to real hardware presents a final step in evaluating the efficacy and performance of the \emph{Hybrid LMC}. We look forward to hardware implementation in future studies. To bridge the sim-to-real gap, we plan on utilizing an environmental encoder, a teacher policy \cite{lee2020learning, kumar2021rma}, and domain randomization techniques. We believe that a LQR (ideally tuned for hardware) within the \emph{Hybrid LMC} will guide the policy in realizing more realistic torques and reduce the risk of undesirable actuation while exploring in the beginning stages of training. 

\subsection{Verifying Hybrid LMC with a Human Data} \label{HMICaseStudy} 
The goal of this experiment was to test the performance of \emph{Hybrid LMC} for locomotion and tracking when given human command signals obtained directly from hardware as a step towards teleoperation (e.g.,HERMES humanoids \cite{wang2015hermes}). This test scenario is important as it suggests viability of using reference trajectories that are irregular and rapidly changing compared to those used in training - $5^{th}$ order velocity polynomials. The recorded human data - body tilt and twist - acquired from the Human Machine Interface \cite{wang2021comparison} is mapped to a desired reference vector $\bm{q}_{des}$ and used for tracking here as seen in Fig. \ref{figMujoco}. Our three main findings were: 1) Compared to the standard LQR with human commands, the \emph{Hybrid LMC} with human commands had a 20\% improvement in position tracking. 2) Compared to the \emph{Hybrid LMC} with preconstructed $5^{th}$ order polynomials, the \emph{Hybrid LMC} with human commands presented slight degradation in position tracking. We believe that because irregular signals - such as those from the human - have high variance, they can negatively affect the RL policy's rate of convergence. 3) The improved \emph{Hybrid LMC} performance was consistent despite being trained using only velocity trajectories.

\begin{table}[t]
\centering
\caption{Verification with different parameters of LQR}
\label{t_ablation_control}
\setlength{\tabcolsep}{0.75pt}
\begin{tabular}{|c|c|c|cc|}
\hline
\multirow{2}{*}{\textbf{}}       & \textbf{LQR parameter}      & \multirow{2}{*}{\textbf{Metric (RMSE)}} & \multicolumn{2}{c|}{\textbf{Method}}                  \\ \cline{2-2} \cline{4-5} 
                                 & \textbf{$K = [K_{xW}, K_{th}, K_{dxW}, K_{dth}]$} &                                         & \multicolumn{1}{c|}{\textbf{LQR}} & \textbf{HybridLC} \\ \hline
\multirow{9}{*}{\textbf{Task 2}} & \multirow{3}{*}{$[-150, -350, -50, -50 ]$}           & Position (m)                   & \multicolumn{1}{c|}{0.1}             &  \textbf{0.058}               \\ \cline{3-5} 
                                 &                             & Velocity (m/s)                & \multicolumn{1}{c|}{0.069}     &   \textbf{0.059}             \\ \cline{3-5} 
                                 &                             & Pitch angle (rad)              & \multicolumn{1}{c|}{0.046}    &  \textbf{0.045}               \\ \cline{2-5} 
                                 & \multirow{3}{*}{$[-50, -200, -20, -20 ]$}           & Position (m)                  & \multicolumn{1}{c|}{\textbf{0.169}}             & 0.204                  \\ \cline{3-5} 
                                 &                             & Velocity (m/s)                 & \multicolumn{1}{c|}{\textbf{0.096}}             &   0.114                 \\ \cline{3-5} 
                                 &                             & Pitch angle (rad)              & \multicolumn{1}{c|}{\textbf{0.047}}             &  0.049                 \\ \cline{2-5} 
                                 & \multirow{3}{*}{$[-25, -100, -10, -10 ]$}           & Position (m)                   & \multicolumn{1}{c|}{\textbf{0.162}}    & 0.519                  \\ \cline{3-5} 
                                 &                             & Velocity (m/s)                 & \multicolumn{1}{c|}{\textbf{0.107}}             & 0.289                   \\ \cline{3-5} 
                                 &                             & Pitch angle (rad)              & \multicolumn{1}{c|}{\textbf{0.048}}             & 0.211                  \\ \hline
\end{tabular}
\end{table}

\subsection{Ablation Study and Analysis of Hybrid LMC} \label{AblationStudy}

\noindent \textbf{Performance comparison of key components:} An ablation study is conducted to investigate and analyze how crucial components of the network affect the performance of \emph{Hybrid LMC}. Based on our results we believe that feeding previous torque commands $\tau^{k-1}_{LQR}, \tau^{k-1}_{\phi}, \tau^{k-1}_c$ and a history of the states $\Theta_{x_e},\Theta_{\dot{x}}$ as an input to \emph{Hybrid LMC} is critical for achieving better performance, as seen in Table \ref{t_ablation_state}. Here, \emph{Hybrid LMC} is trained with the velocity trajectory, Fig. \ref{reward_fig}(c). Each component has a considerable impact on the convergence of the network and in achieving better performance. We hypothesize that leveraging the history state and the previous torque brings benefits to end-to-end (state-to-torque) learning in the broad residual RL framework, also utilized in \emph{Hybrid LMC}.

\noindent \textbf{Verifying Hybrid LMC with varied LQR gains:} 
Here we test the dependency of \emph{Hybrid LMC} on the LQR. The \emph{Hybrid LMC} was trained using the original gains (described in section \ref{result}) but tested using varied LQR gains. From row 1 of Table \ref{t_ablation_control}, we see that increasing LQR gains results in similar performance and superiority of the \emph{Hybrid LMC}. However, decreasing the gains generally resulted in an increase in RMSE error and lead to worse performance. This suggests that high performance of \emph{Hybrid LMC} is dependant on the tuning and response of the feedback controller during training. In other words, switching the gains or feedback controller in deployment is not recommended. We believe that varying the gains randomly during training may help address this issue \cite{kumar2021rma}.


\section{CONCLUSION}
\label{concl}
In this paper, we propose a hybrid learning and model-based controller, \emph{Hybrid LMC}, that combines the strength of a conventional model-based LQR and deep reinforcement learning for more robust tracking in the presence of model uncertainty and parameter changes. Moreover, the ensemble deep reinforcement approach can augment the performance of a standard controller while reducing the variance of a single stochastic-based RL policy. In this manner we are able to perform end-to-end learning directly. By incorporating the ensemble deep RL and the LQR controller, LQR guides the RL policy in generating more appropriate torque within a bounded range, and results in an increase in the sampling efficiency. The ablation studies were conducted and analyzed carefully, to offer a proper method of using \emph{Hybrid LMC}. In all experiments, \emph{Hybrid LMC} outperforms the previous methods and demonstrates generalized performance improvement in the presence of model changes and irregular desired trajectories.\\
\indent In future works, we will apply \emph{Hybrid LMC} on hardware to a wheeled humanoid robot system, SATYRR, to verify the performance of \emph{Hybrid LMC} in the physical world. Based on our result, we expect that \emph{Hybrid LMC} provides an efficient way to train the real system safely while enhancing the performance.   

\section*{Acknowledgements}
The authors are grateful to Youngwoo Sim and Guillermo Colin for their support in designing the figure and discussion for the paper. 


\addtolength{\textheight}{-12cm}   

\bibliographystyle{unsrt}
\bibliography{main.bib}

\end{document}